\def\year{2019}\relax
\documentclass[letterpaper]{article} 
\usepackage{aaai19}  
\usepackage{times}  
\usepackage{helvet}  
\usepackage{courier}  
\usepackage{url}  
\usepackage{graphicx}  
\usepackage{subfigure}
\usepackage{amsmath}
\usepackage{booktabs}
\usepackage{multirow}
\usepackage{multicol}
\begin{document}

\title{Semantic Relationships Guided Representation Learning for Facial Action Unit Recognition}
\author{Guanbin Li\textsuperscript{\rm 1}, Xin Zhu\textsuperscript{\rm 1}, Yirui Zeng\textsuperscript{\rm 1}, Qing Wang\textsuperscript{\rm 1,\rm 2}, Liang Lin\textsuperscript{\rm 1,\rm 2}\thanks{Corresponding author is Liang Lin (Email: linliang@ieee.org). This work was supported in part by the National Key Research and Development Program of China under Grant No.2018YFC0830103, in part by the NSFC-Shenzhen Robotics Projects~(U1613211), in part by the National Natural Science Foundation of China under Grant No.61702565, in part by National High Level Talents Special Support Plan (Ten Thousand Talents Program), in part by the Fundamental Research Funds for the Central Universities under Grant 18lgpy63, and in part by the Science and Technology Plan of Guangdong Province under Grant No.2015B010128009.}\\
\textsuperscript{\rm 1}School of Data and Computer Science, Sun Yat-sen University, China \\
\textsuperscript{\rm 2}Dark Matter AI Inc.\\
liguanbin@mail.sysu.edu.cn, zhux33@mail2.sysu.edu.cn\\ zengyr5@mail2.sysu.edu.cn, wangq79@mail.sysu.edu.cn, linliang@ieee.org
}
\maketitle
\begin{abstract}
Facial action unit (AU) recognition is a crucial task for facial expressions analysis and has attracted extensive attention in the field of artificial intelligence and computer vision. Existing works have either focused on designing or learning complex regional feature representations, or delved into various types of AU relationship modeling. Albeit with varying degrees of progress, it is still arduous for existing methods  to handle complex situations. In this paper, we investigate how to integrate the semantic relationship propagation between AUs in a deep neural network framework to enhance the feature representation of facial regions, and propose an AU semantic relationship embedded representation learning~(SRERL) framework. Specifically, by analyzing the symbiosis and mutual exclusion of AUs in various facial expressions, we organize the facial AUs in the form of structured knowledge-graph and integrate a Gated Graph Neural Network~(GGNN) in a multi-scale CNN framework to propagate node information through the graph for generating enhanced AU representation. As the learned feature involves both the appearance characteristics and the AU relationship reasoning, the proposed model is more robust and can cope with more challenging cases, e.g., illumination change and partial occlusion. Extensive experiments on the two public benchmarks demonstrate that our method outperforms the previous work and achieves state of the art performance.

\end{abstract}

\begin{figure}[t]
\begin{center}
   \includegraphics[width=0.92\columnwidth]{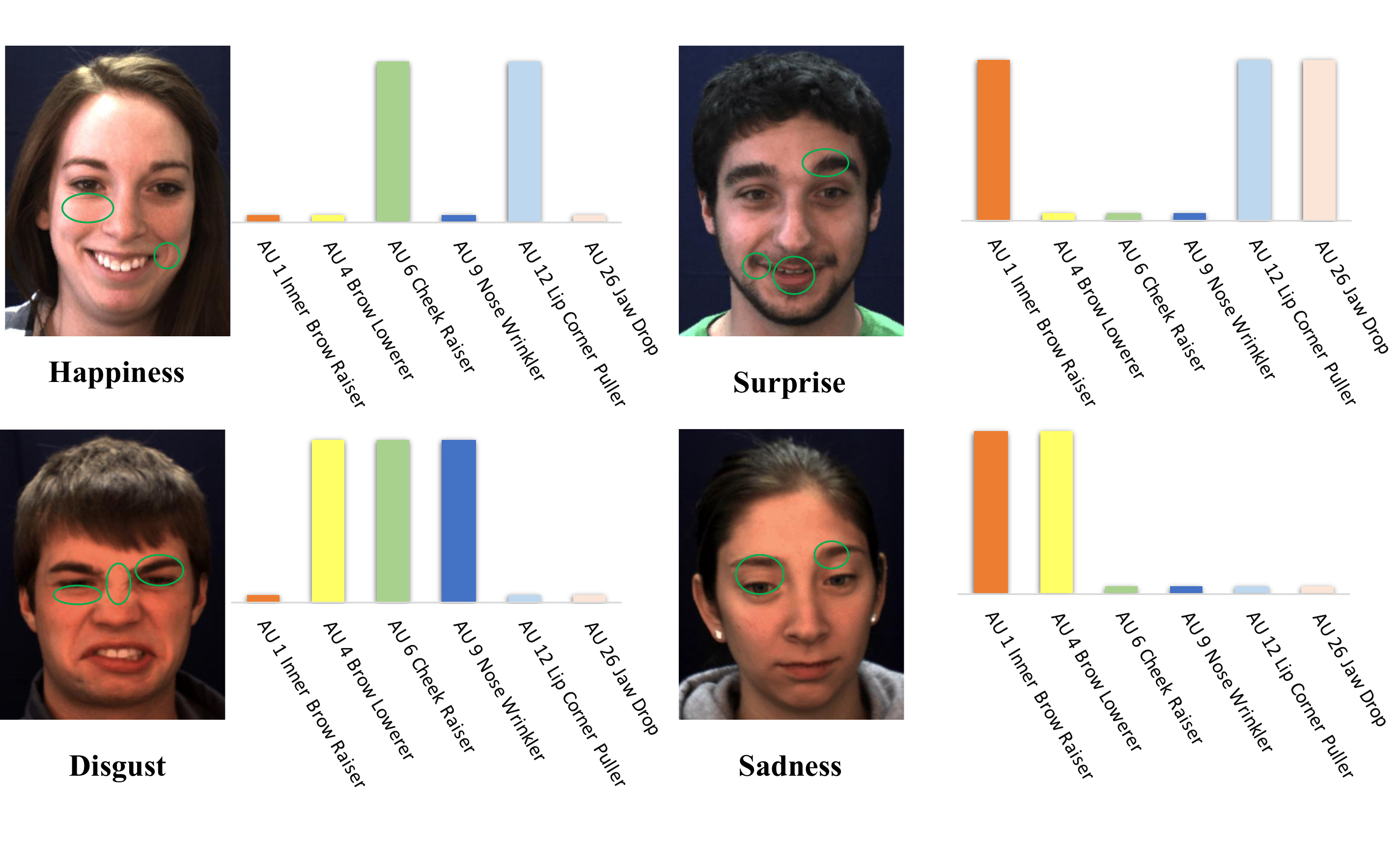}
\end{center}

   \caption{Coupling effect of multiple AUs caused by a variety of facial expressions.}
\label{fig:teaser}
\end{figure}

\section{Introduction}
Facial expression and its behavior is one of the most crucial channels for emotional communication between individuals. Intelligent facial expression analysis has recently attracted wide research interest due to its potential applications in the field of human-robot interaction and computer vision. As defined in the Facial Action Coding System (FACS)~\cite{friesen1978facial}, which is the most widely used and versatile method for measuring and describing facial behaviors, the facial action units (AUs) refer to the local facial muscle actions and plays a paramount role in expressing comprehensive facial expressions. AU recognition has thus become a long-standing research topic in the field of artificial intelligence and computer vision~\cite{li2017eac,li2017action,zhao2016deep}. 

Facial action unit recognition has been mainly treated as a multi-label classification problem that is independent of each other. Most of the traditional methods focus on the design of more discriminative hand-crafted features~(e.g. shape or appearance features) or more effective discriminative learning methods~\cite{valstar2006fully,zhao2015joint,jiang2011action}. In recent years, deep convolutional neural networks have been widely used in AU recognition due to their powerful feature representation and end-to-end efficient learning scheme, and have greatly promoted the development of this field~\cite{zhao2016deep,li2017action,bishay2017fusing,chu2016modeling}. However, recent efforts based on deep convolutional neural networks are indulged in designing deeper and more complex network structures without exception, learning more robust feature representations in a data-driven manner without explicitly considering and modeling the local characteristics of the facial organs and the linkage relationship between the facial muscles. As a common sense, facial action units are not independent from one another, considering the linkage of a same facial expression to multiple action units and the anatomical characteristics of faces. As shown in Fig.~\ref{fig:teaser}, a happiness expression activates both AU6 and AU12, as we generally ``raise cheek'' and ``pull lip corners'' when we smile. On the other hand, due to structural limitations of facial anatomy, some action units are generally not likely to be activated simultaneously. For instance, we can not simultaneously stretch our mouth and raise our cheek. 

Base on the above concerns, some of the research works propose to improve the AU recognition based on the AU relationship modeling. For example, Tong et al.~\cite{tong2007facial} proposed to apply a dynamic Bayesian network (DBN) to model the relationships among different AUs, Wang et al.~\cite{wang2013capturing} developed a more complex model based on restricted Boltzmann machine~(RBM) to exploit the global relationships among AUs and Chu et al.~\cite{ginosar2015century} recently proposed to model both the spatial and temporal prior cues for more accurate facial AU detection. However, existing AU relationship based recognition methods have the following three shortcomings. First of all, existing AU relationship modeling are designed based on handcrafted low level features and are applied as a post-processing or embedded as a priori rule into complex classification models, usually in isolation with the feature representation, and are hence limited to the performance of feature extraction. Secondly, existing methods only attempt to capture local pair-wise AU dependencies based on limited facial expressions observation. This pair-wise data is not combined to form a graph structure for more comprehensive AU relationship reasoning. Last but not the least, since the AU relationship modeling relies on the preamble feature extraction, the whole algorithm framework can not be run end-to-end, which greatly restricts the model efficiency and the performance of the method. Moreover, existing end-to-end deep learning based models also have not explicitly considered the AU relationship in their model design.   

Taking into account the above shortcomings and being inspired by the differentiability of the graph neural network and its superior performance in the entity relationships learning, we  propose an AU semantic relationship embedded representation learning (SRERL) framework to gradually enhance the regional facial feature by fully exploiting the structural collaboration between AUs. Specifically, we organize the facial action units in the form of structured knowledge-graph and propose to integrate a Gated Graph Neural Network (GGNN)~\cite{li2015gated} in a multi-scale CNN framework to propagate node information through the graph. As the proposed network framework is able to embed the globally structured relationships between AUs in the learning process of features in an end-to-end fashion, the learned feature thus involves both the appearance characteristics and the AU relationship reasoning and is more robust towards more challenging cases, e.g., illumination change and partial occlusion. 

In summary, this paper has the following contributions:
\begin{itemize}
\item This work formulates a novel AU semantic relationship embedded representation learning framework which incorporate AU knowledge-graph as an extra guidance for the enhancement of facial region representation. To the best of our knowledge, we are the first to employ graph neural network for AU relationship modeling.
\item
With the guidance of AU knowledge-graph and the differentiable attributes of the gated graph neural network, we are able to collaboratively learn the appearance representation of each facial region and the structured semantic propagation between AUs in an end-to-end fashion. 

\item
We conduct extensive experiments on the widely used BP4D and DISFA datasets and demonstrate the superiority of the proposed SRERL framework over the state-of-the-art facial AU recognition methods. 
\end{itemize}

\section{Related Work}

 Automatic facial action unit recognition has attracted widespread research interest and achieved great progress in recent years. The existing AU recognition methods can be roughly divided into robust feature  representation based methods and semantic relationship modeling based methods.

Traditional feature representation based AU recognition methods focus on the design of more discriminative hand-crafted features. For instance, Valstar et al \cite{valstar2006fully} proposed to analyze the temporal behavior of action units in video and used individual feature GentleBoost templates built from Gabor wavelet features for SVM based AU classification. Baltrusaitis et al~\cite{baltruvsaitis2015cross} designed a facial AU intensity estimation and occurrence detection system based on the fusion of appearance and geometry features. However, these hand-crafted low level feature based algorithms are relatively fragile and difficult to deal with various types of complex situations. In recent years, deep convolutional neural networks have been widely used in a variety of computer vision tasks and have achieved unprecedented progress~\cite{he2016deep,li2018contrast,li2017instance,liu2018facial}. There are also attempts to apply deep CNN to facial AU recognition~\cite{zhao2016deep,li2017action,bishay2017fusing,chu2016modeling}. Zhao et al~\cite{zhao2016deep} proposed a unified architecture for facial AU detection which incorporates a deep regional feature learning and multi-label learning modules. Li et al \cite{li2017action} proposed an ROI framework for AU detection by cropping CNN feature maps with facial landmarks information. However, non of these methods explicitly take into consideration the linkage relationship between different AUs. 

\begin{figure*}[t] 

\begin{center}
	\includegraphics[width=1.80\columnwidth]{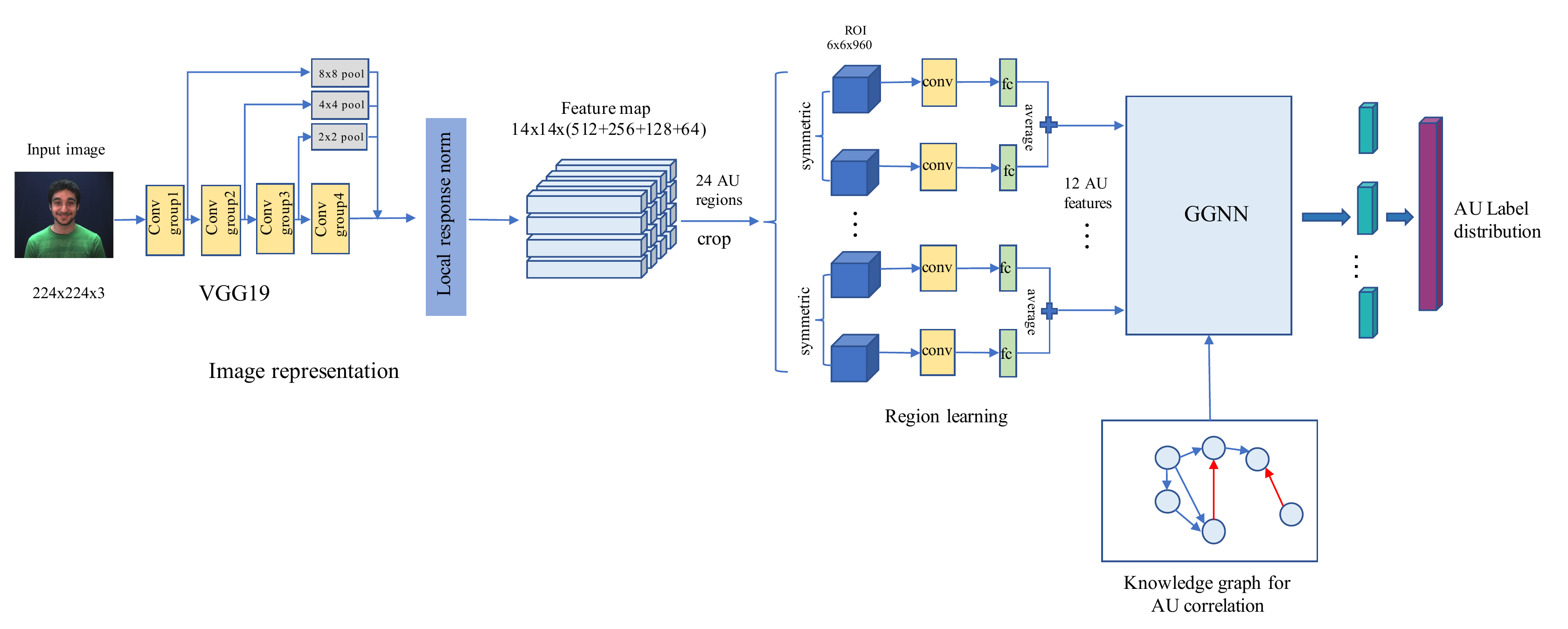}  
\end{center}  
\caption{An overall pipeline of our proposed AU semantic relationship embedded representation learning~(SRERL) framework. It is composed of a multiscale feature learning and cropping module followed by a GGNN for node information propagation. 
}          
\label{fig:framework} 
\vspace*{-7pt}                                                    
\end{figure*}

Considering the linkage effect of facial expressions on various AUs and the anatomical attribute of facial regions, there are research works which rely on action unit relationship modeling to help improve recognition accuracy. Tong et al.~\cite{tong2007facial} proposed a dynamic Bayesian network (DBN) to model relationships between AUs. Wang et al.~\cite{wang2013capturing} further developed a three-layer Restrict Boltzmann Machine~(RBM) to exploit the global relationships between AUs. However, these early works simply model the AU relations from target labels and are independent with feature representation. Wu et al.~\cite{wu2016multiple} proposed a 4-layer RBM to simultaneously capture both feature level and label level dependencies for AU detection. As they are based on handcrafted low level features, the whole algorithm framework can not be performed end-to-end, which greatly restricts the model efficiency and the performance of the method. Recently, Corneanu et al.~\cite{corneanu2018deep} proposed a deep structured inference network (DSIN) for AU recognition which used deep learning to extract image features and structure inference to capture AU relations by passing information between predictions in an explicit way. However, the relationship inference part of DSIN also works as a post-processing step at label level and is isolated with the feature representation.

\section{Method}
In this section, we introduce our semantic relationship embedded representation learning framework in detail. Firstly, we briefly review the gated graph neural network~(GGNN) and introduce the construction of the AU relationship graph, which will work as a guided knowledge-graph for GGNN propagation in our framework. Then, we present the structure of our proposed SRERL framework. The overall pipeline of our framework is illustrated in Figure~\ref{fig:framework}.

\subsection{Preliminary: Gated Graph Neural Network~\cite{li2015gated}}
Given a graph with $N$ nodes and the initial representation of each node, the idea of GGNN is to generate an output for each graph node or a global output, by learning a propagation model similar to LSTM. Specifically, denote the node set as $V$ and its corresponding adjacency matrix $A$. For each node $v \in V$, we define a hidden state representation $h^{t}_{v}$ at every time step $t$. When $t=0$, the hidden state representation is set as the initial node feature $x_{v}$. Adjacency matrix $A$ encodes the structure of the graph. The basic recurrent process of GGNN is formulated as
\begin{equation}
h^{(1)}_{v} = [x^{T}_{v}, 0 ] \label{equation:init}
\end{equation}
\begin{equation}
a^{(t)}_{v} = A^T_v [h^{(t-1)}_{1} \cdots h^{(t-1)}_{|V|}] ^{T} + b \label{equation:update1}
\end{equation}

\begin{equation}\label{equ:3}
z^{t}_{v} = \sigma(W^{z} a^{(t)}_{v}+U^{z} h^{(t-1)}_{v})
\end{equation}

\begin{equation}
r^{t}_{v} = \sigma(W^{r} a^{(t)}_{v}+U^{r} h^{(t-1)}_{v})
\end{equation}

\begin{equation}
\widetilde{h^{t}_{v}} = tanh(Wa^{(t)}_{v} + U(r^{t}_{v} \odot h^{(t-1)}_{v}))
\end{equation}

\begin{equation}\label{equation:update2}
h^{(t)}_{v} = (1-z^{t}_{v}) \odot h^{(t-1)}_{v} + z^{t}_{v} \odot \widetilde{h^{t}_{v}} 
\end{equation}
where $A_{v}$ is a sub-matrix of $A$ which refers to the adjacency matrix for node $v$. $W$ and $U$ are learned parameters. $sigma$ and $tanh$ are the logistic sigmoid and hyperbolic tangent functions. $\odot$ means element-wise multiplication operation. The essence of the iterative update of GGNN evolution is to compute the hidden state of next time-step by looking into the combination of the current hidden state of each node and its adjacent information, as shown in Eq~(\ref{equ:3}-\ref{equation:update2}).

After $T$ time steps, we get the final hidden states. Then the node-level output can be computed as follow
\begin{equation}
o^{v} = g(h^{(T)}_{v}, x_{v}) \label{equation:output}
\end{equation}
where $x_{v}$ is the annotation of node $v$, function $g$ is a fully connected network.
\subsection{Preliminary: AU Relationship Graph Construction}
AU relationship graph represents the globally structured relationships between AUs, which is composed of a node set $V$  and an edge set $E$. Specifically, each node $v\in V$  represents a specific AU and the edges represent the correlations between AUs in our graph. The graph is constructed based on the AU relationship gathered from the training set and some supplementary a priori connections (from common facial expressions and facial anatomy analysis). We detail the construction of $V$ and $E$ as follow,

\begin{figure}[t]
    \centerline{
   \includegraphics[width=0.45\columnwidth]{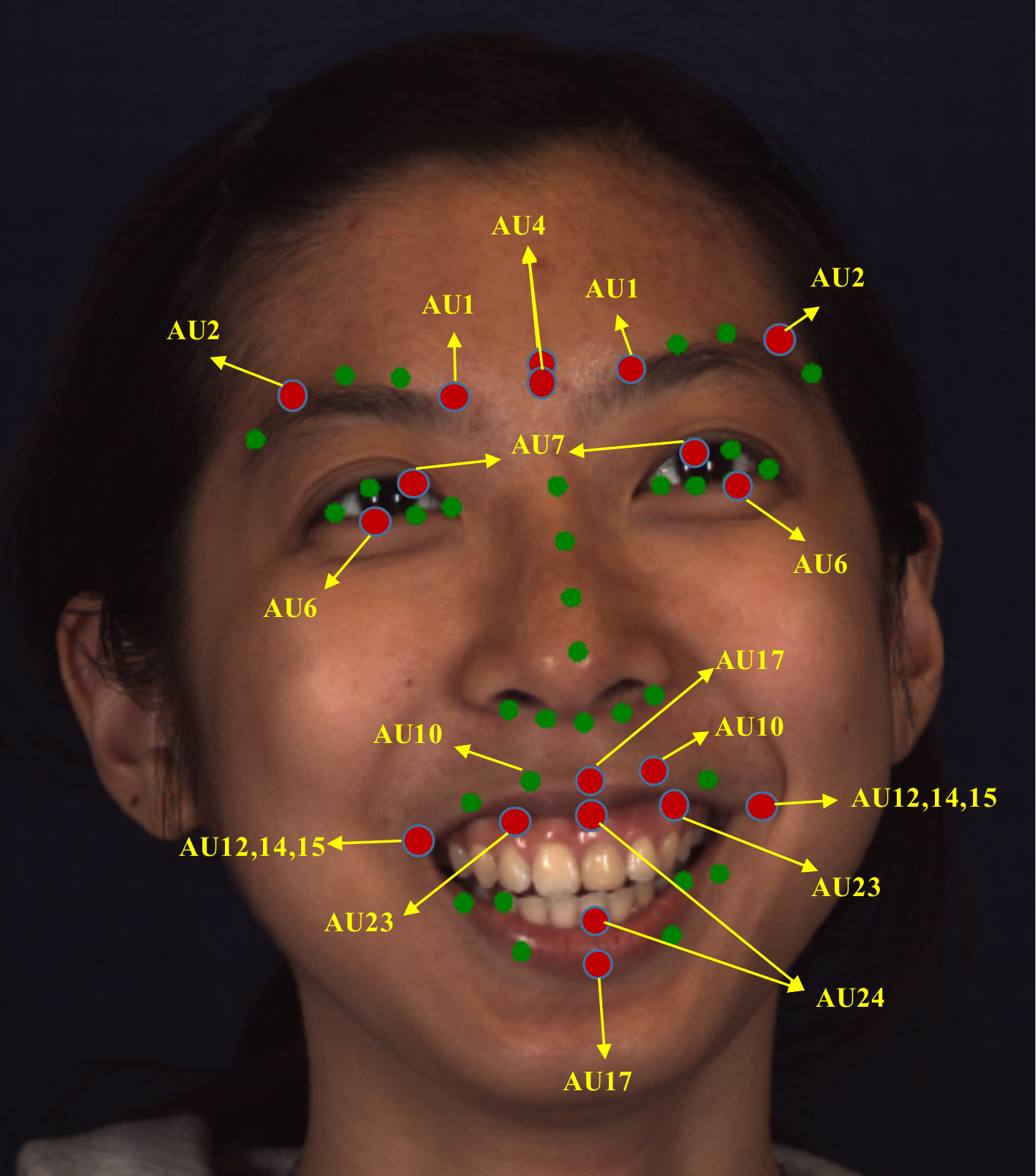} \hfill
    \includegraphics[width=0.45\columnwidth]{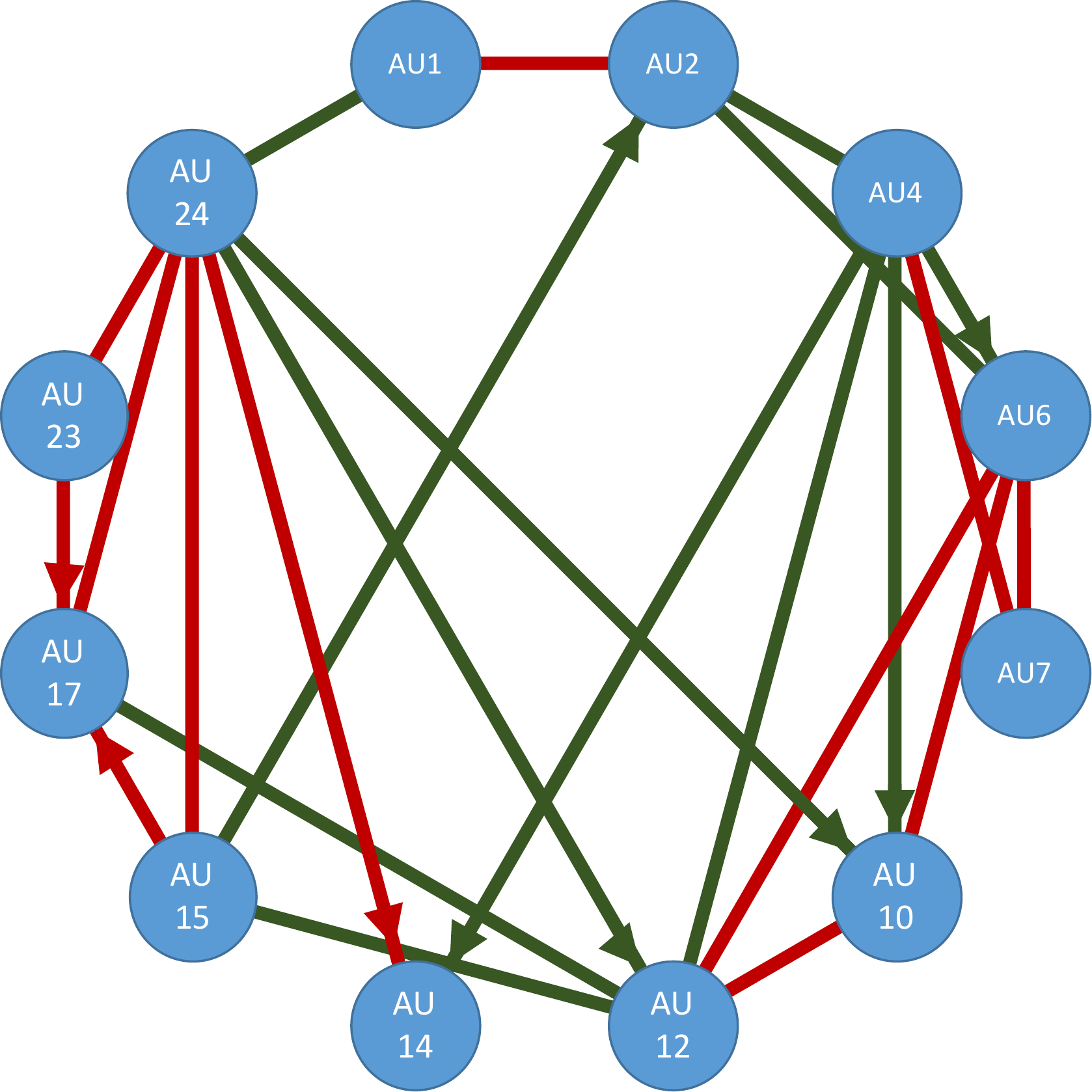}
  }

  \caption{The location of facial AUs and the constructed AU relationship graph.}
  \label{fig:knowledge-graph}
\end{figure}

\subsubsection{Node Set $V$:} Each node in graph represents a corresponding AU. Given a dataset that covers $C$ AUs, the constructed graph is formed by a node set $V$ containing $C$ entities.

\subsubsection{Edge Set $E$:} According to FACS \cite{friesen1978facial}, there exists strong relationship between AUs due to the linkage of each facial expression to multiple action units and the anatomical characteristics of faces. The relationship can further be divided into positive and negative ones, where positive correlation means that two AUs are likely to co-occur while negative correlation refers to that the two rarely appear together. We capture most of the AU relationship  by calculating the conditional probability on the training samples, which is formulated as
\begin{equation}
\begin{split}
a\_\text{pos}_{i,j} = [P(y_{i}=1 | y_{j}=1)- P(y_{i}=1) > p_{\text{pos}}], \label{equation:A_p}
\end{split}
\end{equation}
\begin{equation}
\begin{split}
a\_\text{neg}_{i,j} = [P(y_{i}=1 | y_{j}=1) - P(y_{i}=1) < p_{\text{neg}}] \label{equation:A_n},
\end{split}
\end{equation}
where $y_{n}$ denote the $n$-th AU label. $p_{\text{pos}}$ and $p_{\text{neg}}$ is the threshold for positive and negative relationship, respectively. $a\_\text{pos}_{i,j} \in A_\text{pos}$ and $a\_\text{neg}_{i,j} \in A_\text{neg}$. Thus, we can generate the adjacency matrix of AU relationship graph with the positive relationship matrix $A_{pos}$ and negative relationship matrix $A_{\text{neg}}$, $i.e.$, $A=[A_\text{pos} -I(A),-A_\text{neg},(A_\text{pos} - I(A))^{T},-A_\text{neg}^{T}]$. $I(A)$ denotes the identity matrix of the same dimension as $A$.

However, due to the limitation of dataset, there are still a few common relationships that are not counted.  Referring to the latest research on AU relations~\cite{zhang2018classifier}, we add some additional relationships. Specifically, we add (AU4, AU7), (AU15, AU24) as positive correlation and (AU2, AU6), (AU12, AU15), (AU12, AU17) as negative correlation for BP4D dataset~\cite{zhang2013high}. These relationship are summarized in Figure~\ref{fig:knowledge-graph}, where green lines represent positive correlation and red lines represent negative ones. For clarity, we use a line with arrow to represent a one-way relationship and a line without arrows to represent a two-way relationship.

\subsection{Relationship-Embedded Representation Learning}
As shown in Fig.~\ref{fig:framework}, our proposed SRERL framework is composed of a multiscale CNN based feature learning and cropping module followed by a GGNN for node information propagation. The GGNN takes as input the cropped facial region feature and propagate on the defined knowledge-graph to learn relationship-embed AU feature representation. We elaborate these two components in the following. 

\subsubsection{Multiscale Feature Learning and Cropping}
We choose the VGG19~\cite{simonyan2014very} model as our backbone network, which is composed of 5 groups of convolutional layers with down-sampling. For the sake of the  trade-off between feature performance and its resolution, we use the first 4 groups of convolutional layers for initial appearance feature extraction. The input to the feature extractor is a $224 \times 224$ RGB facial image with its landmarks information. Since the size of the relevant regions of each AU varies~(e.g. the area corresponding to AU6 ``Cheek Raiser'' is larger than that of AU12 ``Lip Corner Puller''), we concatenate the output feature maps of all the four groups~(all resized to $14\times 14$) to form a multiscale global feature. The concatenated feature maps are further fed into a local response normalization layer which could normalize pixels between channels and help to converge. The formulation is given as 

\begin{equation}
b_{c} = \frac{a_{c}}{(k + \frac{\alpha \times C}{n} \sum^{min(C-1, c+n/2)}_{c^{'}=max(0,c-n/2)} a^{2}_{c^{'}})^{\beta}}, \label{equation:lrn}
\end{equation}
where $k$, $\alpha$, $\beta$ are hyperparameters, $C$ represents the channel number of the feature maps. $a_{c}$ denotes the pixel in $c^{th}$ channel and $n$ is set to $2$ which refers to the number of neighboring channels chosen to normalize the pixel $a_{c}$. Then we crop out each AU region according to facial landmarks, which is further fed into separate region learning stream for each local AU feature representation. Specifically, we capture adaptive AU region by resorting to facial landmarks. Since there exists relationship between facial anatomy region and AUs, we can generate the location of each AU center, which is further corresponded to the closest facial landmark. After getting the representative landmark of each facial AU, we crop a $6 \times 6$ region from the concatenated global feature maps for its initial regional feature. Figure \ref{fig:knowledge-graph} illustrates the correspondence between facial AUs and facial landmarks.

We further refer to \cite{li2017action} and adopt separate filters in each cropped AU region to train its representation separately. Suppose we have $C$ AUs. Due to the symmetry of the face, we obtain $2C$ patch-wise feature maps~( each AU corresponds to two patches), and design $2C$ independent regional feature learning branches. For each individual region learning branch, we use a $3\times3$ convolutional layer and a fully connected layer to learn the specific local representation. These convolutional layers and fully connected layers are trained separately, which could help to avoid information interference between different AU regions caused by varies scales of receptive fields.

\subsection{Balanced Loss Function}
Data imbalance is a common problem in AU recognition, especially during multi-label training. However, due to the ensemble training setting of multiple classification units, it is hard to apply effective undersampling or oversampling strategies to balance the data bias. In this work, we design an adaptive loss function for imbalanced data training.

\begin{equation}
\begin{split}
loss = -\frac{1}{C \cdot N} \sum_{j=1}^{N} \sum_{i=1}^{C} 2  \{ r^{i}_{\text{neg}} [l^{i}_{j}=1] \cdot \\
 log(\frac{p^{i}_{j}+0.05}{1.05}) + r^{i}_{\text{pos}}[l^{i}_{j}=0] log(\frac{1.05-p^{i}_{j}}{1.05}) \} \label{equation:loss_cnn}
\end{split}
\end{equation}

\begin{equation}
r^{i}_{\text{pos}} = \frac{\sum_{k=1}^{M}[l^{i}_{k}=1]}{M}
\end{equation}

\begin{equation}
r^{i}_{\text{neg}} = 1-r^{i}_{\text{pos}}
\end{equation}
where $l$ is the ground truth and $p$ is the probability predicted by our method. $[x]$ is an indicator function which returns $1$ when the statement $x$ is true, and $0$ otherwise. Let $C$ denote the number of AUs, and $N$ denote the batch size. $r^{i}_{\text{pos}}$ and $r^{i}_{\text{neg}}$ are constant, which indicate the proportion of positive samples or negative samples in the training set. $M$ denotes the number of samples in the training set.

\subsubsection{GGNN for Action Unit Detection}
As shown in Figure~\ref{fig:framework}, we generate the initial local feature of $2C$ regions using our tailored multiscale feature learning and cropping module. Let $f_{i}$ denotes the feature of a specific AU region, $i \in [0,1,2,…,2C-1]$. Considering the symmetry of a human face, each AU corresponds to two specific regions. The initial feature of each node in GGNN is thus given as
\begin{equation}
x_{v}= \frac{f_{2v} \oplus f_{2v+1}}{2}
\end{equation}
where $\oplus$ denotes element-wise summation. According to Equation \ref{equation:init}, the hidden state vector is initialized with the corresponding feature $x_{v}$. At each time-step $t$, the hidden state can be updated according to Eq~\ref{equation:update1}$-$\ref{equation:update2}. Finally, the output of each node in GGNN can be computed using Eq~\ref{equation:output}, which is further concatenated for final label prediction.

\section{Experimental Results}
\subsection{Datasets}
We evaluated our model on two spontaneous dataset: BP4D\cite{zhang2013high} and DISFA\cite{mavadati2013disfa}.
BP4D contains 2D and 3D facial expression data of 41 young adult subjects, including 18 male and 23 female. Each subject participated in 8 tasks, each of which corresponds to a specific expression. There are totally 146,847 face images with labeled AUs. We refer to~\cite{zhao2016deep} and split the dataset into 3 folds. In which, we take turns to use two folds for training and the other for testing, and report the average results of multiple tests. DISFA contains stereo videos of 27 subjects with different ethnicity. There are totally 130,815 frame images, each of which is labeled with intensities from 0-5. We choose frames with AU intensities higher or equal to C-level as positive samples, and the rest as negative samples. $C$ is chosen as $2$ in our experiment. As with BP4D, we also split the dataset into 3 folds for reliable testing. We use VGG19 model trained on BP4D to directly extract initial appearance feature, and finetune the parameters of separate region learning module and GGNN in DISFA.

\subsection{Evaluation Criteria}
We evaluate our method with two metrics, including F1 score and AUC. F1 score takes into consideration both the precision $p$ and the recall $r$ , and is widely used in binary classification. F1 score can be computed as $F1=2\frac{p \cdot r}{p+r}$, and AUC refers to the area under the ROC curve.

\subsection{Implementation Details}
During the knowledge-graph construction, we set $p_\text{pos}$ as $0.2$ and $p_\text{neg}$ as $-0.03$ in Eq~(\ref{equation:A_p}$-$\ref{equation:A_n}), and set $\alpha=0.002$, $\beta=0.75$, $k=2$ in Eq~\ref{equation:lrn}. We use an Adam optimizer with learning rate of 0.0001 and mini-batch size 64 with early stopping to train our models. For F1-score, we set the threshold of prediction to 0.5. All models are trained using NVIDIA GeForce GTX TITAN X GPU based on the open-source Pytorch platform~\cite{paszke2017automatic}.

\subsection{Ablation Studies} \label{result}

\begin{table*}
\centering

\resizebox{1.65\columnwidth}{!}
{
\begin{tabular}{c|cccccccccccc}
\toprule
AU&1&2&4&6&7&10&12&14&15&17&23&24\\
\midrule
positive samples ratio&0.21&0.17&0.20&0.46&0.55&0.59&0.56&0.47&0.17&0.34&0.17&0.15\\
\bottomrule
\end{tabular}
}
\caption{Data distribution of BP4D dataset}
\label{tab:ratio}
\end{table*}

\begin{table*}
\centering

\resizebox{1.65\columnwidth}{!}
{
\begin{tabular}{c|ccccc|ccccc}
\toprule
\multirow{2}{*}{AU} & \multicolumn{5}{c}{F1-score} & \multicolumn{5}{c}{AUC}\cr
\cmidrule(lr){2-6} \cmidrule(lr){7-11}
&VGG&VGG\_BL&SS\_RL&MS\_RL&SRERL&VGG&VGG\_BL&SS\_RL&MS\_RL&SRERL\cr
\midrule
1&  37.3& 41.6& \textbf{47.1}& 46.6& [46.9] &61.4 &63.1 &[67.5] &66.9 &\textbf{67.6}\cr
2&  32.7& 41.4& 42.9& [44.3]& \textbf{45.3} &59.8 &65.4 &67.9 &[68.6] &\textbf{70.0}\cr
4&  52.3& 44.1& 51.5& [53.1]& \textbf{55.6} &70.6 &65.5 &70.8 &[72.1] &\textbf{73.4}\cr
6&  76.2& 76.4& 76.7& \textbf{77.3}& [77.1] &\textbf{78.5} &78.2 &78.2 &[78.4] &[78.4]\cr
7&  75.7& 73.3& [77.2]& 76.8& \textbf{78.4} &74.1 &73.5 &[75.8] &75.2 &\textbf{76.1}\cr
10& 82.3& 81.2& 82.7& \textbf{83.8}& [83.5] &78.9 &78.8 &[80.4] &\textbf{80.9} &80.0\cr
12& 86.9& 86.0& [87.1]& 86.8& \textbf{87.6} &84.8 &85.5 &[85.7] &85.2 &\textbf{85.9}\cr
14& 55.9& \textbf{64.0}& 62.3& 61.9& [63.9] &61.7 &\textbf{64.5} &64.3 &[64.4] &[64.4]\cr
15& 37.2& 46.7& 49.4& [51.1]& \textbf{52.2} &61.7 &70.2 &71.2 &[73.1] &\textbf{75.1}\cr
17& 57.2& 61.7& 61.6& [63.7]& \textbf{63.9} &68.2 &70.1 &70.3 &[71.4] &\textbf{71.7}\cr
23& 29.4& 40.7& [46.2]& 45.6& \textbf{47.1} &58.5 &66.6 &70.9 &[71.1] &\textbf{71.6}\cr
24& 41.7& 51.6& \textbf{53.8}& 53.2& [53.3] &64.7 &\textbf{76.4} &75.3 &[76.0] &74.6\cr
\midrule
Avg&55.4& 59.1& 61.5& [62.0]& \textbf{62.9} &68.6 &71.5 &73.2 &[73.6] &\textbf{74.1}\cr
\bottomrule
\end{tabular}
}
\caption{Ablation study on the BP4D dataset. It demonstrates the effectiveness of balanced loss design, multiscale CNN as well as relationship-embed feature representation}
\label{tab:ablations_bp4d}
\end{table*}

\begin{table}
\center
\resizebox{0.90\columnwidth}{!}
{
\begin{tabular}{c|cc|cc}
\toprule
\multirow{2}{*}{AU} & \multicolumn{2}{c}{F1-score} & \multicolumn{2}{c}{AUC}\cr
\cmidrule(lr){2-3} \cmidrule(lr){4-5}
&MS\_RL&SRERL&MS\_RL&SRERL\cr
\midrule
1&  39.7& \textbf{45.7} &\textbf{76.4} &76.2\cr
2&  44.5& \textbf{47.8} &80.8 &\textbf{80.9}\cr
4&  52.4& \textbf{59.6} &74.9 &\textbf{79.1}\cr
6&  44.7& \textbf{47.1} &77.8 &\textbf{80.4}\cr
9&  44.7& \textbf{45.6} &\textbf{76.8} &76.5\cr
12& 69.0& \textbf{73.5} &87.7 &\textbf{87.9}\cr
25& \textbf{86.3}& 84.3 &\textbf{91.9} &90.9\cr
26& 42.7& \textbf{43.6} &\textbf{75.3} &73.4\cr
\midrule
Avg&53.0& \textbf{55.9} &80.2 &\textbf{80.7}\cr
\bottomrule
\end{tabular}
}
\caption{The effectiveness of relationship-embedded feature enhancement on DISFA dataset.}
\label{tab:ablations_disfa}
\end{table}

\subsubsection{Effectiveness of Multiscale CNN}
To verify the effectiveness of multiscale feature learning and cropping, we remove the GGNN part of our framework and directly perform multi-label classification on the extracted regional features, we call this variant multiscale representation learning framework~(MS\_RL). We have also implemented a single-scale version for comparison~(SS\_RL). We trained the two variants with the same setting, using our proposed balanced loss function. We have also listed the result of the original VGG network and that trained using balanced loss function~(VGG\_BL) for comparison. As shown in Table~\ref{tab:ablations_bp4d}, SS\_RL achieved higher F1\_score and AUC in most of AUs when compared with VGG\_BL, especially in AU1, 4, 23. Overall, it outperforms VGG\_BL by 2.4\% and 1.7\% in terms of average F1-score and AUC on the BP4D dataset, respectively. We believe that the reason for its higher performance lies in the application of independent regional feature representation, which avoids the feature interference in different AU regions. When compared with SS\_RL, MS\_RL gains 0.50\% and 0.40\% performance improvement w.r.t F1-score and AUC, which evidences the effectiveness of the multiscale setting.

\subsubsection{Effectiveness of Knowledge-Graph based Feature Embedding} 
To verify the effectiveness of AU relationship-embedded feature enhancement, we have compared the performance of our proposed SRERL to those without relationship modeling~(MS\_RL). As shown in Table~\ref{tab:ablations_bp4d} and Table~\ref{tab:ablations_disfa}, we can clearly see the effectiveness. In BP4D dataset, SRERL achieves 0.90\% and 0.50\% performance boost w.r.t F1-score and AUC when compared with MS\_RL. The accuracy improvement is more significant in DISFA dataset, which is 2.90\% and 0.50\% in terms of F1-score and AUC respectively. At the same time, it can be seen from the analysis of individual AU performance that relationship modeling actually plays a crucial role in improving the accuracy of recognition. For instance, there exists strong positive relationship between AU1 and AU2. In BP4D dataset, AU1 and AU2 increased by 0.3\% and 1.0\% in F1-score, and increased by 0.7\% and 1.4\% in AUC respectively when compared to MS\_RL. Meanwhile, SRERL achieved 6.0\% and 3.3\% higher F1-score in DISFA dataset for AU1 and AU2. On the other side, AU12 and AU15 are negatively correlated. In BP4D dataset, AU12 and AU15 increased by 0.8\% and 1.1\% in F1-scroce, and increased by 0.7\% and 2.0\% in AUC. The experimental results well demonstrate that the relationship-embedded representation learning can greatly enhance the regional facial feature by fully exploiting the structural collaboration between AUs.

\subsubsection{Effectiveness of Balance Loss}
Table~\ref{tab:ratio} illustrates the data distribution of the BP4D data set. We can clearly see that the occurrence rate of AU2, 15, 23, 24 are less than 20\% while that of AU10 is close to 60\%. From Table~\ref{tab:ablations_bp4d}, we can find that the AUs which are of higher occurrence rate achieve higher performance in VGG method. For example, the occurrence rate of AU12 is 56\% in the dataset, and the F1-score and AUC of AU12 are 86.9\% and 84.8\% respectively. While the occurrence rate of AU23 is 17\%, its corresponding accuracy is relatively lower, which is 29.4\% and 58.5\% respectively.

As shown in table \ref{tab:ablations_bp4d}, VGG with balance loss~(VGG\_BL) outperforms the original VGG on many AUs, especially in those with lower occurrence rate. For example, it increases AU2 and AU15 by 8.7\% and 9.5\% w.r.t F1-score, and 5.6\% and 8.5\% in terms of AUC, respectively. Compared to original VGG, VGG\_BL achieved about 4\% and 3\% higher F1-score and AUC in average.

\begin{table*}[t]
\center

\resizebox{1.75\columnwidth}{!}
{
\begin{tabular}{c|ccccccc|ccccc}
\toprule
\multirow{2}{*}{AU} & \multicolumn{7}{c}{F1-score} & \multicolumn{5}{c}{AUC}\cr
\cmidrule(lr){2-8} \cmidrule(lr){9-13}
&LSVM &JPML &LCN &DRML &ROI &DSIN &Ours &LSVM &JPML &LCN &DRML &Ours\cr
\midrule{}
1&  23.2& 32.6& 45.0& 36.4 &36.2 &\textbf{51.7} &[46.9] &20.7 &40.7 &51.9 &[55.7] &\textbf{67.6}\cr
2&  22.8& 25.6& 41.2& [41.8] &31.6 &40.4 &\textbf{45.3} &17.7 &42.1 &50.9 &[54.5] &\textbf{70.0}\cr
4&  23.1& 37.4& 42.3& 43.0 &43.4 &\textbf{56.0} &[55.6] &22.9 &46.2 &53.6 &[58.8] &\textbf{73.4}\cr
6&  27.2& 42.3& 58.6& 55.0 &\textbf{77.1} &76.1 &\textbf{77.1} &20.3 &40.0 &53.2 &[56.6] &\textbf{78.4}\cr
7&  47.1& 50.5& 52.8& 67.0 &[73.7] &73.5 &\textbf{78.4} &44.8 &50.0 &[63.7] &61.0 &\textbf{76.1}\cr
10& 77.2& 72.2& 54.0& 66.3 &\textbf{85.0} &79.9 &[83.5] &73.4 &[75.2] &62.4 &53.6 &\textbf{80.0}\cr
12& 63.7& 74.1& 54.7& 65.8 &[87.0] &85.4 &\textbf{87.6} &55.3 &60.5 &[61.6] &60.8 &\textbf{85.9}\cr
14& [64.3]& \textbf{65.7}& 59.9& 54.1 &62.6 &62.7 &63.9 &46.8 &53.6 &[58.8] &57.0 &\textbf{64.4}\cr
15& 18.4& 38.1& 36.1& 33.2 &[45.7] &37.3 &\textbf{52.2} &18.3 &50.1 &49.9 &[56.2] &\textbf{75.1}\cr
17& 33.0& 40.0& 46.6& 48.0 &58.0 &[62.9] &\textbf{63.9} &36.4 &42.5 &48.4 &[50.0] &\textbf{71.7}\cr
23& 19.4& 30.4& 33.2& 31.7 &38.3 &[38.8] &\textbf{47.1} &19.2 &51.9 &50.3 &[53.9] &\textbf{71.6}\cr
24& 20.7& [42.3]& 35.3& 30.0 &37.4 &41.6 &\textbf{53.3} &11.7 &53.2 &47.7 &[53.9] &\textbf{74.6}\cr
\midrule
Avg&35.3& 45.9& 46.6& 48.3 &56.4 &[58.9] &\textbf{62.9} &32.2 &50.5 &54.4 &[56.0] &\textbf{74.1}\cr
\bottomrule
\end{tabular}
}
\caption{Comparison of quantitative results on the BP4D dataset. Our proposed SRERL achieves the best performance, which outperforms the second best method by 4.0\% and 18.1\% in terms of F1-score and AUC respectively.}
\label{tab:compare_bp4d}
\end{table*}

\begin{table*}[t]
\centering
\resizebox{1.75\columnwidth}{!}
{
\begin{tabular}{c|cccccc|cccc}
\toprule
\multirow{2}{*}{AU} & \multicolumn{6}{c}{F1-score} & \multicolumn{4}{c}{AUC}\cr
\cmidrule(lr){2-7} \cmidrule(lr){8-11}
&LSVM &LCN &DRML &ROI &DSIN &Ours &LSVM &LCN &DRML &Ours\cr
\midrule
1&  10.8& 12.8& 17.3& 41.5& [42.4] &\textbf{45.7} &21.6 &44.1 &[53.3] &\textbf{76.2}\cr
2&  10.0& 12.0& 17.7& 26.4& [39.0] &\textbf{47.8} &15.8 &52.4 &[53.2] &\textbf{80.9}\cr
4&  21.8& 29.7& 37.4& [66.4]& \textbf{68.4} &59.6 &17.2 &47.7 &[60.0] &\textbf{79.1}\cr
6&  15.7& 23.1& 29.0& \textbf{50.7}& 28.6 &[47.1] &8.7  &39.7 &[54.9] &\textbf{80.4}\cr
9&  11.5& 12.4& 10.7& 8.5& \textbf{46.8} &[45.6] &15.0 &40.2 &[51.5] &\textbf{76.5}\cr
12& 70.4& 26.4& 37.7& \textbf{89.3}& 70.8 &[73.5] &\textbf{93.8} &54.7 &54.6 &[87.9]\cr
25& 12.0& 46.2& 38.5& [88.9]& \textbf{90.4} &84.3 &3.4  &[48.6] &45.6 &\textbf{90.9}\cr
26& 22.1& 30.0& 20.1& 15.6& [42.2] &\textbf{43.6} &20.1 &[47.0] &45.3 &\textbf{73.4}\cr
\midrule
Avg&21.8& 24.0& 26.7& 48.5& [53.6] &\textbf{55.9} &27.5 &46.8 &[52.3] &\textbf{80.7}\cr
\bottomrule
\end{tabular}
}
\caption{Comparison of quantitative results on the DISFA dataset. Our proposed SRERL achieves the best performance, which outperforms the second best method by 2.3\% and 28.4\% in terms of F1-score and AUC respectively.}
\label{tab:compare_disfa}
\end{table*}

\subsection{Comparison with the State of the Art}
We compare our method to alternative methods, including linear SVM(LSVM) \cite{fan2008liblinear}, Joint Patch and Multi-label Learning(JPML) \cite{zhao2015joint}, ConvNet with locally connected layer(LCN) \cite{taigman2014deepface}, Deep Region and Multi-label Learning(DRML) \cite{zhao2016deep}, Region Adaptation, Multi-label Learning(ROI) \cite{li2017action} and Deep Structure Inference Network(DSIN) \cite{corneanu2018deep}. 

Table \ref{tab:compare_bp4d} and table \ref{tab:compare_disfa} show the result of 12 AUs on BP4D and 8 AUs on DISFA. We can clearly witness that our model outperform all of these state-of-the-art methods. Compared to LSVM, our method achieved 27.6\% and 41.9\% higher F1-score and AUC on BP4D dataset, and achieved 34.1\% and 53.2\% higher F1-score and AUC on DISFA dataset. The reason for the higher performance is that LSVM only use handcraft feature and independent label inference. Our method outperforms JPML by 17\% in terms of F1-score and 23.6\% in AUC on BP4D in average. The main reason lies in that JPML use predefined feature and is not end-to-end trainable. LCN, DRML, ROI and DSIN are four state of the art methods which use convolutional neural network for end-to-end feature learning and multi-label classification. Compared to LCN, our method reaches higher F1-frame and AUC in all AUs on BP4D and DISFA dataset. DRML proposed to divide feature map into 8*8 patches and use region learning for each patch. ROI used facial landmarks to crop the feature and maps to 20 patch, which also applied region learning for each patch. Compared with DRML and ROI, our model achieves 14.6\% and 6.5\% higher performance in F1-score on BP4D while achieves 29.2\% and 7.4\% higher w.r.t F1-score on DISFA. More impressively, our proposed SRERL greatly outperforms DSIN by 4\% and 2.3\% in terms of F1-score on BP4D and DISFA respectively. Noted that DSIN also simultaneously modeled the deep feature learning and the structured AU relationship in an unified framework. However, the relationship inference part of DSIN works as a post-processing step at label level and is being isolated with the feature representation. Compared with DSIN, our performance gain well proves the significance of the joint optimization of feature expression and semantic relationship modeling for face action unit classification, and on the other hand evidences the effectiveness of graph neural network in exploiting more thorough relationship in facial AU representation.

\section{Conclusion}
In this paper, we have proposed a semantic relationship embedded representation learning framework to well incorporate our constructed AU relationship knowledge-graph as extra guidance information for end-to-end feature representation learning in a deep learning framework. Specifically, the SRERL framework is composed of a multiscale feature learning and cropping module followed by a gated graph neural network for node feature propagation across the defined knowledge-graph. To the best of our knowledge, we are the first to explore the AU relationship based on a differentiable graph neural network module in a feature representation level. Experimental results on the widely used BP4D and DISFA datasets have demonstrated the superiority of the proposed SRERL framework over the state-of-the-art methods.

\bibliographystyle{aaai} 
\bibliography{srgrl}

\end{document}